\documentclass[10pt,twocolumn,letterpaper]{article}

\usepackage[pagenumbers]{cvpr} % To force page numbers, e.g. for an arXiv version

\usepackage{array}
\usepackage{booktabs}
\usepackage{multirow}
\usepackage{makecell}
\newcommand{\tabincell}[2]{\begin{tabular}{@{}#1@{}}#2\end{tabular}}

\definecolor{cvprblue}{rgb}{0.21,0.49,0.74}
\usepackage[pagebackref,breaklinks,colorlinks,allcolors=cvprblue]{hyperref}

\def\paperID{7109} % *** Enter the Paper ID here
\def\confName{CVPR}
\def\confYear{2025}

\title{Point Cloud Unsupervised Pre-training via 3D Gaussian Splatting}

\author{Hao Liu\textsuperscript{1} \quad Minglin Chen\textsuperscript{2} \quad Yanni Ma\textsuperscript{2} \quad Haihong Xiao\textsuperscript{3} \quad Ying He\textsuperscript{1}\\
\textsuperscript{1}Nanyang Technological University \quad \textsuperscript{2}Sun Yat-Sen University \\ \textsuperscript{3}South China University of Technology\\
}

\begin{document}
\maketitle
\begin{abstract}
Pre-training on large-scale unlabeled datasets contribute to the model achieving powerful performance on 3D vision tasks, especially when annotations are limited. However, existing rendering-based self-supervised frameworks are computationally demanding and memory-intensive during pre-training due to the inherent nature of volume rendering. In this paper, we propose an efficient framework named \textbf{GS$^3$} to learn point cloud representation, which seamlessly integrates fast 3D Gaussian Splatting into the rendering-based framework. The core idea behind our framework is to pre-train the point cloud encoder by comparing rendered RGB images with real RGB images, as only Gaussian points enriched with learned rich geometric and appearance information can produce high-quality renderings. Specifically, we back-project the input RGB-D images into 3D space and use a point cloud encoder to extract point-wise features. Then, we predict 3D Gaussian points of the scene from the learned point cloud features and uses a tile-based rasterizer for image rendering. Finally, the pre-trained point cloud encoder can be fine-tuned to adapt to various downstream 3D tasks, including high-level perception tasks such as 3D segmentation and detection, as well as low-level tasks such as 3D scene reconstruction. Extensive experiments on downstream tasks demonstrate the strong transferability of the pre-trained point cloud encoder and the effectiveness of our self-supervised learning framework. In addition, our GS$^3$ framework is highly efficient, achieving approximately 9$\times$ pre-training speedup and less than 0.25$\times$ memory cost compared to the previous rendering-based framework \textit{Ponder}.
\end{abstract}    
\section{Introduction}
\label{sec:intro}

In recent years, we have witnessed the tremendous success of deep neural networks using supervised learning across various vision tasks such as object detection. However, acquiring large amounts of high-quality and diverse annotations is expensive and time-consuming, especially for 3D annotations. For example, labeling an indoor scene consisting of thousands of 3D points requires approximately 30 minutes \cite{ScanNet}. In this context, self-supervised learning (SSL) has emerged as a viable alternative to supervised learning for tasks with limited annotations. 

\begin{figure}
    \centering
    \includegraphics[width=\columnwidth]{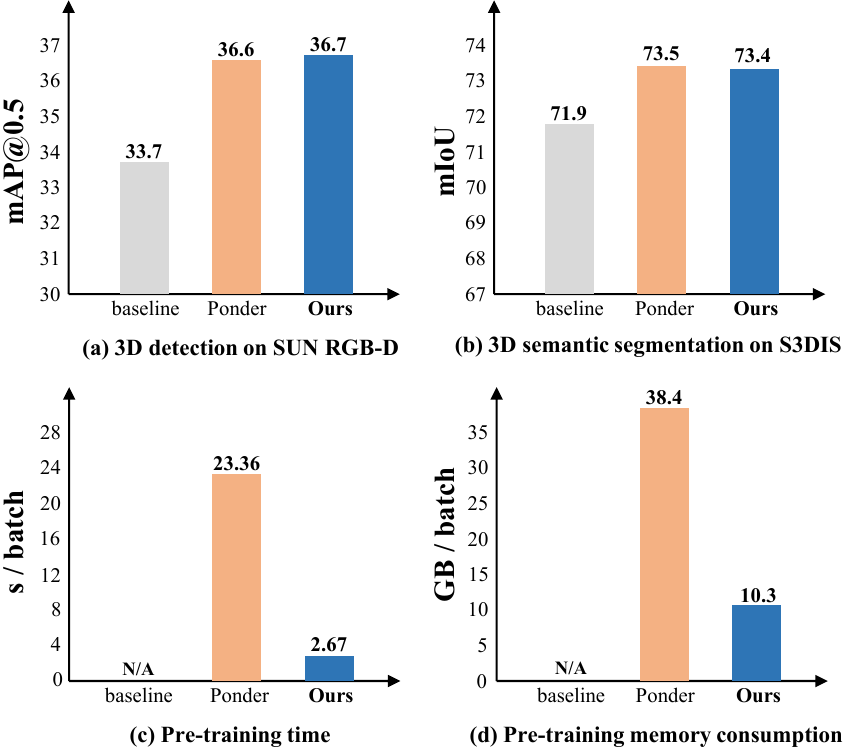}
    \caption{Comparison of 3D detection performance \textit{mAP@0.5}, 3D segmentation accuracy \textit{mIoU}, pre-training time and memory consumption of Ponder \cite{Ponder} and our GS$^3$. The pre-training time and memory usage of our method are measured at a rendered image resolution of 320 $\times$ 240. Due to limited computational resources, the pre-training time of Ponder with 76,800 sampling rays is estimated based on its result with 4,800 rays. Memory consumption for pre-training is reported only for Ponder with 4,800 rays.}
    \label{fig_1}
\end{figure}
Existing SSL methods for 3D point clouds are broadly grouped into three categories: completion-based, contrast-based and rendering-based. Completion-based methods \cite{PointMAE,PointM2AE,VoxelMAE,GD-MAE} typically design a pretext task to reconstruct masked point clouds from incomplete observations, drawing inspiration from the masked autoencoder (MAE) \cite{MAE}. Despite remarkable progress, this paradigm remains highly challenging and under-explored due to the irregular and sparse nature of point clouds. Furthermore, such methods are sensitive to the masking rate and the selection of missing parts. Contrast-based methods \cite{PointContrast,DepthContrast,STRL,4DContrast,wu2023masked} are designed to learn invariant representations under different geometric transformations. However, these methods converge slowly and rely heavily on elaborate strategies such as positive/negative sampling and data augmentation.

Subsequently, Huang et al. \cite{Ponder} proposed a novel rendering-based framework named \textit{Ponder}, which back-projects multi-view RGB-D images into 3D space to build a 3D feature volume and renders the images via differentiable volume rendering. The model is pre-trained by minimizing the difference between the rendered image and the input image. Although the learned features can effectively encode the scene's geometry and appearance cues, this method not only requires dense multi-view images as input and depth maps as addition supervision, but also demands substantial memory and computational resources due to the dozens of point queries along each ray.

Motivated by this, we propose an efficient 3D \textbf{G}aussian \textbf{S}platting-based \textbf{S}elf-\textbf{S}upervised (\textbf{GS$^3$}) framework that accepts sparse view RGB-D images. The proposed GS$^3$ formulates a 3D Gaussian Splatting (GS)-based neural rendering pretext task, which leverages point cloud features to produce scene 3D Gaussians and adopts a fast tile-based rasterizer to render the RGB images. Thanks to real-time rendering framework 3D GS, our model significantly reduces the computational burden and memory costs during pre-training compared to \textit{Ponder} \cite{Ponder}, as shown in Fig. \ref{fig_1}. Furthermore, to render high-quality novel view images, 3D GS enforces the point cloud encoder to capture rich geometry and appearance information, which further facilitates the pre-training of the point cloud encoder. To the best of our knowledge, our framework is the first attempt to explore generalizable 3D GS for point cloud self-supervised learning. Specifically, we first lift the input sparse view RGB-D images to 3D space to generate a group of colored point clouds. Then, the generated point clouds are input into a point cloud encoder to extract point-wise features, which are used to predict the point-aligned Gaussian locations and primitive parameters. Finally, given specific camera intrinsic parameters and poses, we employ a real-time tile-based renderer to produce RGB images. Our model is trained by minimizing the difference between the rendered and input RGB images. The point cloud encoder pre-trained by our SSL framework can serve as a strong initialization for various downstream tasks, including 3D semantic segmentation, 3D instance segmentation, 3D object detection and 3D scene reconstruction. In summary, main contributions of our paper are listed as follows:

\begin{itemize}
    \item We propose a 3D Gaussian Splatting-based self-supervised model, which seamlessly integrates generalizable 3D GS into the rendering-based SSL framework.
    \item The proposed model, GS$^3$, is capable of accommodating various point cloud encoder. The encoder pre-trained by our framework can be effectively transferred to various downstream tasks.
    \item Extensive experiments on four downstream tasks show the excellent transferability of the pre-trained encoders, thus validating the effectiveness of our framework. In addition, our framework achieves 9$\times$ pre-training speedup and less than 0.25$\times$ memory cost compared to \textit{Ponder}.
\end{itemize}
\section{Related works}
\label{sec:related_works}

\subsection{Self-supervised learning in 3D point clouds}

Self-supervised learning (SSL) is a label-free approach where a model learns effective representations by designing and solving pretext an unsupervised task. Existing methods for 3D point clouds are roughly divided into \textbf{completion-based}, \textbf{contrast-based} and \textbf{rendering-based}.

\textbf{Completion-based methods} \cite{PointMAE,PointM2AE,MaskPoint,VoxelMAE,GD-MAE,GeoMAE,Occupancy-MAE} typically devise a pretext task to reconstruct missing point clouds from partial or incomplete observations. PointMAE \cite{PointMAE} introduces a transformer-based autoencoder that reconstructs masked point patches by optimizing a set-to-set Chamfer distance \cite{CD}. PointM2AE \cite{PointM2AE} introduces a multi-scale strategy for hierarchical point cloud encoding and reconstruction. In MaskPoint \cite{MaskPoint}, Liu et al. designed a pretext task for binary classification to distinguish between masked and unmasked points. However, these methods are constrained to indoor scenes. Hess et al. \cite{VoxelMAE} proposed Voxel-MAE, which leverages voxel representations to facilitate MAE pre-training on large-scale outdoor point clouds. Subsequently, Yang et al. \cite{GD-MAE} proposed a sparse pyramid transformer to extract multi-scale features from pillar-shaped point clouds, and then used a generative decoder to unify feature scales and recover masked feature markers. 

\textbf{Contrast-based methods} \cite{PointContrast,DepthContrast,hou2021exploring,STRL,4DContrast,jiang2021guided,wu2023masked} are designed to learn robust representations under different geometric transformations. Xie et al. \cite{PointContrast} learned invariant representations by computing correspondences between two different views of the same point cloud scene. Zhang et al. \cite{DepthContrast} utilized various input representations, such as voxels and points, allowing the framework to handle arbitrary 3D data. Subsequent works have been proposed to enhance feature representations by leveraging spatio-temporal cues in 4D sequence data \cite{STRL,4DContrast} or by developing new augmentation strategies to produce hard positive/negative pairs \cite{wu2023masked}. For example, Chen et al. \cite{4DContrast} synthesized static 3D scene data with moving objects to create 4D sequence data and thus establish temporal correspondences. Wu et al. \cite{wu2023masked} proposed a combination of spatial and photometric augmentations to generate diverse training pairs.

In addition to the above two categories, Huang et al. \cite{Ponder,PonderV2} first proposed a novel \textbf{rendering-based} framework \textit{Ponder}. It back-projects the input RGB-D images into 3D space and employs a point cloud encoder to extract features for each point. These point features are organized into a 3D feature volume, which is then used to render RGB images and depth maps via volume rendering. These rendered images are compared with the input RGB-D images for supervision. Subsequent works, UniPad \cite{UniPad} and PRED \cite{PRED}, apply volumetric rendering to outdoor point cloud SSL. However, a key hurdle of this framework lies in its high computational and memory demands, inherent to volume rendering. In this paper, we propose a computationally efficient framework for self-supervised point cloud learning using 3D Gaussian Splatting.

\subsection{Neural scene representation}
Neural scene representation aims to model the geometry and appearance of 3D scenes using neural networks. Neural radiance field (NeRF) \cite{NeRF} is one of the representative methods, which represents scenes through simple multi-layer perceptrons (MLPs) and renders scene RGB images via volume rendering. Building on this framework, several works \cite{Mip-NeRF,NeuS,NeuralUDF,InstantNGP,TensoRF,VolSDF} propose new ray sampling strategies to accelerate rendering and incorporate SDF \cite{DeepSDF} or UDF \cite{GeoUDF} representations to enhance the quality of rendered images. Despite significant progress, these methods are constrained by high computational demands, largely due to the numerous point queries required per ray during rendering.

Recently, Kerbl et al. \cite{3DGS} proposed a novel neural scene representation, 3D Gaussian Splatting (3DGS), which models scenes using a set of anisotropic Gaussian points and employs a tile-based rasterizer for image rendering. This approach achieves impressive real-time rendering speeds while maintaining high-quality novel view synthesis. However, 3DGS-based methods \cite{Scaffold-gs,Compact3DGS,Mip-splatting} require scene-specific optimization. To this end, Charatan et al. \cite{pixelSplat} proposed pixelSplat, the first generalizable Gaussian model that directly predicts pixel-aligned Gaussian primitive parameters in a feed-forward manner. Chen et al. \cite{MVSplat} constructed a lightweight cost volume to replace the epipolar transformer in pixelSplat for cross-image encoding. Wang et al. \cite{FreeSplat} proposed an adaptive cost view aggregation module and a pixel-wise triplet fusion strategy to enable free-view synthesis over across a wide range of views. Our work is inspired by recent advances in generalizable 3DGS.

\subsection{3D scene understanding}
According to the processing of input point clouds, existing network architectures for 3D scene understanding can be broadly classified into projection-based \cite{MV3D,ContFuse}, discretization-based \cite{graham20183d,VoxelNet,PointPillars,CenterTube} and point-based methods \cite{liu2020semantic,PointRCNN,3DSSD,AnchorPoint}. Projection-based methods project the point clouds into 2D space and then adopt well-established 2D scene understanding pipelines. Discretization-based methods partition 3D space into regular cells to facilitate the operation of 3D convolutions. The major drawback of these methods is that their efficiency and accuracy are highly correlated to cell resolution. The advent of sparse convolutions (SpConvs) achieves an optimal trade-off between efficiency and accuracy. Point-based methods directly consume raw point cloud data, but are limited in large-scale scenes due to heavy computational burden and high memory costs. In this work, we pre-train the point-based PointNet++ \cite{PointNet++} and discretization-based Sparse Residual U-Net (SR-UNet) \cite{choy20194d} implemented with SpConv \cite{SpConv}. 
\section{Methodology}
\begin{figure*}[t]
    \centering
    \includegraphics[width=\linewidth]{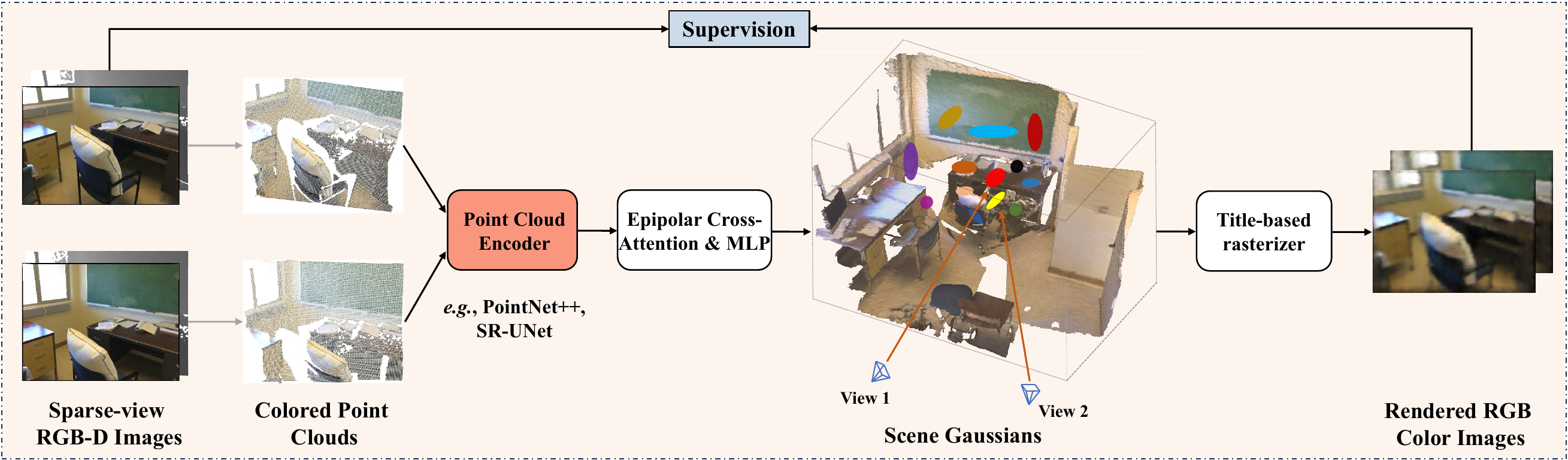}
    \caption{The overall framework of the proposed GS$^3$. Given sparse-view RGB-D images, we back-project them into 3D space to generate colored point clouds. A point cloud encoder is then used to extract point-wise features, which are used to predict scene Gaussians in a point-aligned manner. These Gaussians are rendered into RGB images through a differentiable tile-based rasterizer. The point cloud encoder is pre-trained by comparing the rendered images with the real images.}
    \label{fig_2}
\end{figure*}

We propose GS$^3$, a \textbf{G}aussian \textbf{S}platting-based \textbf{S}elf-\textbf{S}upervised learning framework for 3D point clouds, as shown in Fig. \ref{fig_2}. First, the input RGB-D images are back-projected into 3D space to form 3D point clouds according to the provided camera intrinsic parameters and poses (Section \ref{Section_3_2}). Next, we use a point cloud encoder to extract point-wise features (Section \ref{Section_3_3}), which are then used to produce scene 3D Gaussians that represent the scene's geometry and appearance, enabling RGB image rendering through a tile-based rasterizer (Section \ref{Section_3_4}). Finally, the rendered images are compared with the input images as a supervision signal for our model (Section \ref{Section_3_5}). The point cloud encoder pre-trained by our framework can be fine-tuned for various downstream tasks.

\subsection{3D point cloud generation}
\label{Section_3_2}
Our method takes as input sparse view RGB-D images $\{\mathbf{I}_i,\mathbf{D}_i|\mathbf{I}_i\in{\mathbb{R}^{H\times{W}\times{3}}},\mathbf{D}_i\in{\mathbb{R}^{H\times{W}}}\}_{i=1}^N$, along with camera intrinsic parameters $\{\mathbf{K}_i\}_{i=1}^N$ and poses $\{\mathbf{T}_i|\mathbf{T}_i=[\mathbf{R}_i|\mathbf{t}_i]\}_{i=1}^N$, where $N$ is the number of input views, $H$ and $W$ are the height and width of the input image, respectively. Each camera pose $\mathbf{T}_i$ is defined by its rotation matrix $\mathbf{R}_i$ and translation vector $\mathbf{t}_i$. Following the pinhole camera model \cite{hartley2003multiple}, we back-project the RGB-D images into 3D space to facilitate the pre-training of the point cloud encoder as follows:
\begin{equation}
\begin{bmatrix}
x \\
y \\
z
\end{bmatrix} = \mathbf{R}^{-1} \left( d \cdot \mathbf{K}^{-1} \begin{bmatrix} u \\ v \\ 1 \end{bmatrix} - \mathbf{t} \right)
\end{equation}
where $[x,y,z]^T$ is the generated 3D point in the world coordinate system, and $d$ is the depth value of pixel $(u,v)$. In addition, to better model the scene appearance, we append the RGB color of each pixel to its corresponding 3D points.

\subsection{3D feature encoder}
\label{Section_3_3}
After generating the 3D point clouds $\mathcal{X}$ from the input RGB-D images, we use a point cloud encoder $f_p$ to extract point-wise features $\mathcal{F}\in{\mathbb{R}^{(H\times{W})\times{C}}}$, \textit{i.e.}, $\mathcal{F}=f_p(\mathcal{X})$, where $(H\times{W})$ is the number of point clouds, $C$ is the feature dimension. In this work, we employ the point-based network PointNet++ \cite{PointNet++} and the discretization-based network SR-UNet \cite{PointContrast} as our feature encoders. Additional details and visualizations of our encoders are provided in the supplementary material.

\subsection{Pre-training with 3D Gaussian Splatting}
\label{Section_3_4}
This section describes how our approach seamlessly incorporates 3D Gaussian Splatting (GS) into the self-supervised learning framework. We begin with a brief overview of 3D GS and then discuss how to produce scene 3D Gaussians from extracted point cloud features. Finally, we utilize a tile-based rasterizer to render RGB images for supervision.

\textbf{Brief introduction to 3D Gaussian Splatting}: 3D GS represents the scene with a dense set of anisotropic 3D Gaussians. Each Gaussian $G(\mathbf{x})$ is parameterized by its center (\textit{i.e.}, mean) $\mathbf{\mu}\in{\mathbb{R}^3}$ and covariance matrix $\mathbf{\Sigma}$:
\begin{equation}
    G(\mathbf{x})=exp(-\frac{1}{2}(\mathbf{x}-\mathbf{\mu})^T\mathbf{\Sigma}^{-1}(\mathbf{x}-\mathbf{\mu}))
\end{equation}
The covariance matrix $\mathbf{\Sigma}$ is decomposed into a scaling matrix $\mathbf{S}$ and a rotation matrix $\mathbf{R}$, \textit{i.e.}, $\mathbf{\Sigma}=\mathbf{R}\mathbf{S}\mathbf{S}^T\mathbf{R}^T$. In addition to $\mathbf{\mu}$ and $\mathbf{\Sigma}$, 3D GS includes additional parameters, such as spherical harmonics (SH) coefficients $\mathbf{c}$, assigned to each Gaussian to better model the view-dependent appearance of the scene. The Gaussians are then projected onto 2D space to produce rendered RGB images:
\begin{equation}\label{eq3}
C(v)=\sum_{i\in{\mathcal{N}}}\mathbf{c}_i\alpha_i\prod_{j=1}^{i-1}(1-\alpha_j)
\end{equation}
where $C(v)$ represents the RGB color value of the rendered image at pixel $v$, and $\mathcal{N}$ denotes the set of all Gaussians that contribute to pixel $v$. The rendered image is compared with the real image to optimize the Gaussian position $\mathbf{\mu}$ and other primitive parameters, \textit{e.g.}, $\mathbf{\Sigma}$, $\alpha$, $\mathbf{c}$. However, it is infeasible to apply vanilla 3D GS for unsupervised pre-training due to the requirement for per-scene optimization.

\textbf{Generating scene 3D Gaussians from point cloud features}: Inspired by generalizable GS \cite{pixelSplat,MVSplat}, we predict scene 3D Gaussians from extracted point cloud features. In other words, we predict one or multiple Gaussians for each point. Taking two input views as an example, i.e., $\mathcal{F}_1$ and $\mathcal{F}_2$, we first use a cost volume module \cite{MVSplat} or epipolar line transformer \cite{FreeSplat} to perform cross-view feature encoding, thus $\hat{\mathcal{F}_1}$ and $\hat{\mathcal{F}_2}$. Then, we use a feed-forward network $f_{\theta}$ to learn a mapping from the encoded point cloud features $\hat{\mathcal{F}}\in{\mathbb{R}^{(H\times{W})\times{C}}}$ to 3D Gaussian parameters:
\begin{equation}\label{GS_formulation}
    f_{\theta}: \left\{ \hat{\mathcal{F}_i} \right\}_{i=1}^N \mapsto k\times \left\{ (\Delta\mathbf{\mu}_j,\mathbf{\Sigma}_j,\mathbf{\alpha}_j,\mathbf{c}_j) \right\}_{j=1}^{(H\times{W})\times{N}}
\end{equation}
\begin{equation}
    \mathbf{\mu}_j=p_i+\Delta\mathbf{\mu}_j
\end{equation}
where $p_i$ is the 3D coordinate of the $j$-th point, $\Delta\mu_j$ is the offset between the $j$-th point and its predicted Gaussian center $\mu_j$. $k$ is the number of Gaussians predicted at each point. In this way, we predict the scene 3D Gaussian parameters from the scene point cloud features in a point-aligned manner, and thus the total number of 3D Gaussian is $k\times(H\times{W})\times{N}$ for $N$-view input RGB-D images with shape $H\times{W}$. Different from vanilla GS, which requires per-scene optimization, our formulation in Eq. \ref{GS_formulation} can optimize multiple scenes simultaneously. This facilitates the integration of 3D GS into self-supervised learning framework.

\textbf{Masked point modeling (MPM)}: Inspired by MAE, we introduce MPM for self-supervised learning of 3D point clouds. Similar to completion-based frameworks \cite{MaskPoint}, we mask 50\% of the point clouds generated by back-projecting the RGB-D images. However, rather than reconstructing the masked point clouds from the remaining points, we use the visible points to predict scene Gaussians and render RGB images. This strategy encourages the point cloud encoder to capture the precise geometric and spatial information of the point clouds, thereby enhancing the ability to understand the complete scene.

\textbf{Differentiable rendering}: After producing the scene's 3D Gaussians, we employ a differentiable tile-based rasterizer to render view-dependent RGB images according to the provided camera poses. Specifically, given a viewpoint with its viewing transformation matrix $W$, we project the 3D Gaussians onto the 2D image plane:
\begin{equation}
    \mu_{2D}=\frac{{J}{W}\mu_{3D}}{Z},
    \Sigma_{2D}=JW\Sigma_{3D}W^TJ^T
\end{equation}
where $Z$ is the depth value of the Gaussian, and $J$ is the Jacobian matrix of the affine approximation of the projective transformation. For each image pixel, we first determine the set of Gaussians that contribute to that pixel, and then calculate the alpha value $\alpha$ of each Gaussian, \textit{i.e.}, $\alpha_i=o_iG_i^{2D}(v)$. Notice that, $o_i$ is the opacity value of the $i$-th Gaussian, $G_i^{2D}(\cdot)$ denotes the function of the projected $i$-th Gaussian. Finally, we multiply the Gaussian colors by their corresponding $\alpha$ values and then accumulate them along the ray direction to obtain the image pixel value, as Eq. \ref{eq3}.

\subsection{Pre-training objectives}
\label{Section_3_5}
Different from Ponder \cite{Ponder}, we only use RGB images as supervision. The total pre-training loss $L$ is the weighted sum of image color loss $L_{color}$ and LPIPS \cite{LPIPS} loss $L_{lpips}$: $L=L_{color}+\lambda \cdot L_{lpips}$.

\textbf{Image color loss $L_{color}$}: It is a traditional pixel-level loss that measures color consistency between rendered and ground-truth pixels. We apply MSE loss for supervision:
\begin{equation}
    L_{color}=\frac{1}{H\times{W}}\sum_{i=1}^{H\times{W}}(I_r(i)-I_{gt}(i))^2
\end{equation}
where $H\times{W}$ is the number of image pixels. $I_r$ and $I_{gt}$ denote the rendered image and the ground-truth image, respectively.

\textbf{LPIPS loss $L_{lpips}$}: It is a perception-based image patch-level similarity metric, which is designed to measure the high-level differences between the render image and the ground-truth image. $L_{lpips}$ is complementary to $L_{color}$, and they are usually optimized together to obtain high-quality rendered images.
\begin{equation}
    L_{lpips}=\sum_{l} \frac{1}{M_l} \sum_{i=1}^{M_l} \| w_l \odot (\hat{f_l}(I_r(i)) - \hat{f_l}(I_{gt}(i))) \|_2^2
\end{equation}
where $M_l=H_l\times{W_l}$, and $\hat{f_l}(I_r)$ denotes the normalized feature map of the rendered image $I_r$ at the $l$-th layer of the VGG \cite{VGG} network. $w_l$ denotes the channel-wise weights. $H_r$ and $W_l$ are the height and width of the feature map at the $l$-th layer.

In our experiments, we follow the loss weight setting of pixelSplat \cite{pixelSplat}, i.e., $\lambda=0.05$.
\section{Experiments}
\label{sec:experiments}

\subsection{Experimental settings}

\subsubsection{Datasets} We use ScanNet v2 \cite{ScanNet} as the pre-training dataset. ScanNet v2 contains a total of 1513 indoor scenes, where 1201 scenes with diverse 3D annotations (e.g., 3D box annotations, point-level and instance-level segmentation annotations) are allocated for training, and 312 scenes are reserved for testing. Each scene comprises hundreds of temporally continuous RGB-D images along with the corresponding camera intrinsic parameters and poses.

\subsubsection{Implementation details} In our self-supervised framework, we take two-view RGB-D images with overlapping regions as input, and back-project them into 3D space to form point cloud data. The resolution of the input images is $320\times{240}$, and the frame interval between input views is 5. We use point-based PointNet++ and discretization-based SR-UNet as point cloud encoders, both of which have 128 output feature dimensions. More details for PointNet++ and SR-UNet are provided in the supplementary material.

We pre-train our model with a batch size of 4 for 100 epochs, where each batch corresponds to one scene. The model is pre-trained on a single NVIDIA A100 40G GPU, and the entire pre-training process takes approximately three days. We use AdamW \cite{AdamW} to optimize model parameters, where the initial learning rate is set to 1e-4 and weight decay is set to 0.05. The cosine annealing \cite{cosine_anneling} strategy is adopted to update the learning rate, where the minimum learning rate is set to 1e-6. To obtain diverse training samples, we apply the same random rotations along the X, Y, and Z axes to both the point cloud data and camera poses. The rotation angle ranges for the X, Y and Z axes are $[-\pi/64,\pi/64]$, $[-\pi/64,\pi/64]$ and $[-\pi,\pi]$, respectively. For fine-tuning of downstream tasks, we use the pre-trained point cloud encoder as initialization and follow their experimental settings.

\begin{table*}[t]
    \centering
    \resizebox{0.85\textwidth}{!}{
    \begin{tabular}{r|ccc|cc}
        \Xhline{2.0\arrayrulewidth}
        \multirow{2}{*}{Method} & \multirow{2}{*}{\tabincell{c}{Detection \\ Model}} & \multirow{2}{*}{\tabincell{c}{Pre-training \\ Type}} & \multirow{2}{*}{\tabincell{c}{Pre-training \\ Epochs}} & \multicolumn{2}{c}{SUN RGB-D} \\
        & & & & mAP@0.5$\uparrow$ & mAP@0.25$\uparrow$ \\
        \Xhline{2.0\arrayrulewidth}
        3DETR \cite{3DETR} & 3DETR & - & - & 30.3 & 58.0 \\
        Point-BERT \cite{PointBERT} & 3DETR & Completion-based & 300 & - & - \\
        MaskPoint \cite{MaskPoint} & 3DETR & Completion-based & 300 & - & - \\
        \Xhline{2.0\arrayrulewidth}
        VoteNet \cite{VoteNet} & VoteNet & - & - & 33.7 & 57.7 \\
        STRL \cite{STRL} & VoteNet & Contrast-based & 100 & 35.0 & 58.2 \\
        RandomRooms \cite{RandomRooms} & VoteNet & Contrast-based & 300 & 35.4 & 59.2 \\
        PointContrast \cite{PointContrast} & VoteNet & Contrast-based & - & 34.8 & 57.5 \\
        PC-FractalDB \cite{PC-FractalDB} & VoteNet & Contrast-based & - & 33.9 & 59.4 \\
        DepthContrast \cite{DepthContrast} & VoteNet & Contrast-based & 1000 & 35.4 & 60.4 \\
        IAE \cite{IAE} & VoteNet & Contrast-based & 1000 & 36.0 & 60.4 \\
        \Xhline{2.0\arrayrulewidth}
        Ponder \cite{Ponder} & VoteNet & Rendering-based & 100 & 36.6 & 61.0 \\
        \textbf{GS$^3$ (Ours)} & VoteNet & Rendering-based & 100 & 36.7 \textcolor{red}{(+3.0)} & 61.3 \textcolor{red}{(+3.6)} \\
        \Xhline{2.0\arrayrulewidth}
    \end{tabular}}
    \caption{Comparative \textbf{3D object detection} results among current self-supervised methods on the SUN RGB-D and ScanNet v2 datasets. The red number in each bracket denotes the performance improvement over the corresponding baseline method.}
    \label{table1}
\end{table*}

\begin{table}[h]
    \centering
    \resizebox{\columnwidth}{!}{
    \begin{tabular}{c|c|cc}
        \Xhline{2.0\arrayrulewidth}
        Method & \tabincell{c}{\#Sampling \\ Rays} & \tabincell{c}{Pre-training Time \\ (s/batch) $\downarrow$} & \tabincell{c}{Pre-training Memory \\ (GB/batch) $\downarrow$} \\
        \Xhline{2.0\arrayrulewidth}
        Ponder$^\dagger$ \cite{Ponder} & 4800 & 1.46 & 38.4 \\
        Ponder$^\dagger$ \cite{Ponder} & 76800 & 23.36 & - \\
        \textbf{GS$^3$ (Ours)} & 76800 & 2.67 & 10.3 \\
        \Xhline{2.0\arrayrulewidth}
    \end{tabular}}
    \caption{Comparison of pre-training time and memory consumption for rendering-based frameworks. All results are obtained on a single NVIDIA A100 40G GPU. $\dagger$ denotes the reproduced results. The pre-training time of Ponder with 76800 sampling rays is estimated from the result of its 4,800 sampling rays. }
    \label{table2}
\end{table}

\begin{table}[h]
    \centering
    \resizebox{0.8\columnwidth}{!}{
    \begin{tabular}{r|cc}
        \Xhline{2.0\arrayrulewidth}
        Method & mAP@0.5$\uparrow$ & mAP@0.25$\uparrow$ \\
        \Xhline{2.0\arrayrulewidth}
         VoteNet \cite{VoteNet} & 33.5 & 58.6 \\
         3DETR \cite{3DETR} & 37.5 & 62.7 \\
         3DETR-m \cite{3DETR} & 47.0 & 65.0 \\
         H3DNet \cite{H3DNet} & 48.1 & 67.2 \\
        \Xhline{2.0\arrayrulewidth}
        Ponder \cite{Ponder} + H3DNet & 50.9 & 68.4 \\
        \textbf{GS$^3$} + H3DNet & 50.4 \textcolor{red}{(+2.3)} & 68.0 \textcolor{red}{(+0.8)} \\
        \Xhline{2.0\arrayrulewidth}
    \end{tabular}}
    \caption{3D object detection results of GS$^3$ with H3DNet on the ScanNet v2 dataset. The red number in each bracket denotes the improvement over the corresponding baseline.}
    \label{table3}
\end{table}

\subsection{Fine-tuning on downstream tasks}

To validate the effectiveness of the proposed GS$^3$ framework, we pre-train the point cloud encoder on the ScanNet v2 dataset and transfer the weights as initialization for downstream tasks.

\subsubsection{High-level tasks}

\textbf{3D object detection.} We use two indoor scene datasets, SUN RGB-D \cite{SUN-RGBD} and ScanNet v2 \cite{ScanNet}, to evaluate the transferability of our pre-trained encoder to the 3D object detection task. SUN RGB-D contains 10,335 indoor scenes, each of which provides RGB-D images, camera poses and 3D box annotations. Following \cite{Ponder}, VoteNet \cite{VoteNet} and H3DNet \cite{H3DNet} are selected as our baselines. We use mean average precision (mAP) as the primary metric, with the IoU thresholds set to 0.25 and 0.5, respectively.

Table \ref{table1} reports the quantitative results among current self-supervised methods on downstream 3D detection task. Table \ref{table2} presents the pre-training time and memory consumption of current rendering-based framework. Notice that, due to limited computational resources, the pre-training overhead of Ponder \cite{Ponder} is measured with 4,800 sampled rays. All pre-training overheads are obtained on a single NVIDIA A100 40G GPU.   We observe that the baseline VoteNet with our GS$^3$ gains remarkable improvements, increasing mAP@0.5 by 3.0\% on the SUN RGB-D datasets. Ponder \cite{Ponder} is a rendering-based framework that leverages NeRF \cite{NeRF} to generate rendered images for pre-training. Our proposed rendering-based framework GS$^3$ achieves comparable improvements to the baseline VoteNet as Ponder. However, our method achieves 9$\times$ pre-training speedup and less than 0.25$\times$ memory consumption compared to Ponder. In addition, compared with recent contrast-based method IAE \cite{IAE}, the point cloud features learned by our method achieve higher mAP values with a gain of 0.7\% on the SUN RGB-D dataset.

To further verify the effectiveness of our GS$^3$ framework, we follow Ponder to combine GS$^3$ with a more powerful baseline method H3DNet \cite{H3DNet}. Table \ref{table3} shows the 3D detection results. We can see that, our method outperforms H3Net by 2.3\% and 0.8\% in terms of mAP@0.5 and mAP@0.25, respectively.

\textbf{3D semantic segmentation.} We use the ScanNet v2 \cite{ScanNet} and S3DIS \cite{S3DIS} datasets to evaluate the semantic segmentation performance of our fine-tuned model. Different from ScanNet v2, which reconstructs 3D scenes from RGB-D images, S3DIS uses LiDAR scanner to capture point clouds in indoor environments. It contains approximately 272 indoor samples from 6 different buildings, with point-wise semantic and instance-level segmentation annotations for each sample. The strong MinkUNet \cite{MinkUNet} is selected as our baseline. We use mean IoU (mIoU) and mean accuracy (mAcc) as the major evaluation metrics.

Table \ref{table4} reports the quantitative results of our GS$^3$ combined with MinkUNet. Our method significantly improves the baseline MinkUNet on both S3DIS and ScanNet v2 datasets, regardless of whether the voxel size is 2cm or 5cm. Specifically, with a voxel size of 2cm, the mIoU is increased by 1.6\% and 1.5\% for S3DIS and ScanNet v2, respectively. Similar improvements are observed in mAcc metric (S3DIS: 1.1\%, ScanNet v2: 0.4\%) as well. Furthermore, we find that our method achieves comparable improvements (73.4\% vs. 73.5\%) over the existing rendering-based approach, Ponder \cite{Ponder}, on the ScanNet v2 dataset. This indicates that our GS$^3$ framework is capable of effectively improving the 3D semantic segmentation performance of the baseline methods.

\begin{table}[t]
    \centering
    \resizebox{\columnwidth}{!}{
    \begin{tabular}{r|cccc}
        \Xhline{2.0\arrayrulewidth}
        \multirow{2}{*}{Method} & \multicolumn{2}{c}{S3DIS (Area-5)} & \multicolumn{2}{c}{ScanNet v2} \\
        & mIoU$\uparrow$ & mAcc$\uparrow$ & mIoU$\uparrow$ & mAcc$\uparrow$ \\
        \Xhline{2.0\arrayrulewidth}
        PointNet \cite{PointNet} & 41.1 & 49.0 & - & -\\
        PointNet++ \cite{PointNet++} & - & - & 53.5 & - \\
        KPConv \cite{KPConv} & 67.1 & 72.8 & 69.2 & - \\
        SparseConvNet \cite{SparseConvNet} & - & - & 69.3 & - \\
        Point Transformer \cite{PT} & 70.4 & 76.5 & 70.6 & - \\
        MinkUNet \cite{MinkUNet} & - & - & 72.2 & - \\
        Ponder + MinkUNet \cite{Ponder} &  - & - & 73.5 & - \\
        \Xhline{2.0\arrayrulewidth}
        MinkUNet$^\dagger$ (5cm) \cite{MinkUNet} & 62.8 & 70.6 & 66.6 & 75.0\\
        \multirow{2}{*}{\textbf{GS$^3$} + MinkUNet (5cm)} & 63.8 & 71.3 & 68.2 & 76.4 \\
        & \textcolor{red}{(+1.0)} & \textcolor{red}{(+0.7)} & \textcolor{red}{(+1.6)} & \textcolor{red}{(+1.4)} \\
        MinkUNet$^\dagger$ (2cm) \cite{MinkUNet} & 68.5 & 75.2 & 71.9 & 80.6 \\
        \multirow{2}{*}{\textbf{GS$^3$} + MinkUNet (2cm)} & 70.1 & 76.3 & 73.4 & 81.0 \\
        & \textcolor{red}{(+1.6)} & \textcolor{red}{(+1.1)} & \textcolor{red}{(+1.5)} & \textcolor{red}{(+0.4)} \\
        \Xhline{2.0\arrayrulewidth}
    \end{tabular}}
    \caption{Comparative \textbf{3D semantic segmentation} results on the S3DIS and ScanNet v2 datasets. $\dagger$ denotes the reproduced results.}
    \label{table4}
\end{table}

\textbf{3D instance segmentation.} We evaluate the transferability of our GS$^3$ to the 3D instance segmentation task on the S3DIS \cite{S3DIS} and ScanNet v2 \cite{ScanNet} datasets. The classic PointGroup \cite{PointGroup} is selected as the baseline. We use average AP  and AP with a IoU threshold of 0.5 as the major evaluation metrics. The average AP is calculated by averaging the AP values across IoU thresholds from 50\% to 95\% with an interval of 5\%. Table \ref{table5} presents the quantitative results of PointGroup with our GS$^3$. We find that the proposed GS$^3$ significantly improves the baseline of PointGroup at different voxel resolutions. Specifically, with a voxel size of 2cm, the average AP and AP@0.5 on the S3DIS dataset are increased by 0.7\% and 1.7\%, respectively. Consistent improvements are also seen on the ScanNet v2 dataset. This demonstrates the effectiveness of our GS$^3$ framework for 3D instance segmentation task.

\begin{table}[t]
    \centering
    \resizebox{\columnwidth}{!}{
    \begin{tabular}{r|cccc}
        \Xhline{2.0\arrayrulewidth}
        \multirow{2}{*}{Method} & \multicolumn{2}{c}{S3DIS (Area-5)} & \multicolumn{2}{c}{ScanNet v2}\\
         & avg. AP$\uparrow$ & AP@0.5$\uparrow$ & avg. AP$\uparrow$ & AP@0.5$\uparrow$ \\
        \Xhline{2.0\arrayrulewidth}
        3D-SIS \cite{3D-SIS} & - & - & - & 18.7 \\
        GSPN \cite{GSPN} & - & - & 19.3 & 37.8\\
        PointGroup \cite{PointGroup} & - & 57.8 & 34.8 & 56.7 \\
        DyCo3D \cite{DyCo3D} & - & - & 35.4 & 57.6 \\
        \Xhline{2.0\arrayrulewidth}
        PointGroup$^\dagger$ (5cm) \cite{PointGroup} & 40.1 & 55.7 & 27.2 & 49.1 \\
        \multirow{2}{*}{\textbf{GS$^3$} + PointGroup (5cm)} & 40.4 & 57.7 & 28.1 & 50.6 \\
        & \textcolor{red}{(+0.3)} & \textcolor{red}{(+2.0)} & \textcolor{red}{(+0.9)} & \textcolor{red}{(+1.5)} \\
        PointGroup$^\dagger$ (2cm) \cite{PointGroup} & 45.2 & 59.4 & 35.2 & 57.6 \\
        \multirow{2}{*}{\textbf{GS$^3$} + PointGroup (2cm)} & 45.9 & 61.1 & 37.0 & 59.2 \\
        & \textcolor{red}{(+0.7)} & \textcolor{red}{(+1.7)} & \textcolor{red}{(+1.8)} & \textcolor{red}{(+1.6)} \\
        \Xhline{2.0\arrayrulewidth}
    \end{tabular}}
    \caption{Comparative \textbf{3D instance segmentation} results on the S3DIS and ScanNet v2 dataset. $\dagger$ denotes the reproduced results.}
    \label{table5}
\end{table}

\subsubsection{Low-level task}
In addition to high-level perception tasks, we evaluate the fine-tuned model on low-level task to further validate the effectiveness of our GS$^3$ framework.

\textbf{3D scene reconstruction.} We select the Synthetic Indoor Scene \cite{ConvONet} dataset to evaluate the scene reconstruction performance of our fine-tuned model. Following \cite{Ponder}, the commonly used ConvONet \cite{ConvONet} is chosen as our baseline. We use volumetric IoU, normal consistency (NC) and F-score with the threshold of 1\% as the main metrics.

Table \ref{table6} shows the quantitative results of the baseline and several self-supervised approaches on downstream 3D scene reconstruction task. Our method achieves competitive results with a volumetric IoU of 79.7\% and a F-score of 91.6, which improves the baseline ConvONet by 1.9\% and 1.0\% in terms of volumetric IoU and F-score, respectively. This shows that our GS$^3$ framework is effective in improving the scene reconstruction performance of the baseline, demonstrating the strong transferability of our GS$^3$. In addition, compared with other self-supervised methods, our rendering-based approach outperforms the contrast-based method IAE \cite{IAE} by 4.0\% in terms of volumetric IoU, while achieving comparable performance to the recent rendering-based framework Ponder \cite{Ponder}.

\begin{table}[h]
    \centering
    \resizebox{\columnwidth}{!}{
    \begin{tabular}{r|cccc}
        \Xhline{2.0\arrayrulewidth}
        Method & Encoder & IoU$\uparrow$ & NC$\uparrow$ & F-Score$\uparrow$ \\
        \Xhline{2.0\arrayrulewidth}
        ConvONet \cite{ConvONet} & PointNet++ & 77.8 & 88.7 & 90.6 \\
        IAE \cite{IAE} & PointNet++ & 75.7 & 88.7 & 91.0 \\
        Ponder \cite{Ponder} & PointNet++ & 80.2 & 89.3 & 92.0 \\
        \textbf{GS$^3$ (Ours)} & PointNet++ & 79.7 \textcolor{red}{(+1.9)} & 89.0 \textcolor{red}{(+0.3)} & 91.6 \textcolor{red}{(+1.0)} \\
        \Xhline{2.0\arrayrulewidth}
    \end{tabular}}
    \caption{Comparative \textbf{3D scene reconstruction} results on the Synthetic Indoor Scene dataset. The red number in each bracket denotes the improment over the corresponding baseline.}
    \label{table6}
\end{table}

\subsection{Ablation study}
In this section, we conduct a group of ablation experiments to justify our framework design and parameter settings. These experiments are performed on the 3D semantic segmentation task, evaluated on the S3DIS \textit{Area-5} set.

\textbf{Mask ratio.} We propose a masked point modeling strategy to augment point cloud data, and thus encourage point cloud encoder to learn contextual features. In this ablation experiment, we investigate the impact of mask ratio on our method. Table \ref{table7} presents the 3D segmentation results of our fine-tuned model with different mask ratios, ranging from 0\% to 90\%. We observe that our method achieves the best results with a mIoU of 70.1\% and a mAcc of 76.3\% when mask ratio is 0.5. This may be because a larger mask ratio retains few Gaussian points, causing the point cloud encoder to not fully learn the geometry and appearance information of the scene, while a smaller value may result in redundant Gaussian points. Overall, our GS$^3$ framework is insensitive to the mask ratio and can improve the baseline method at different mask ratios. 

\begin{table}[h]
    \centering
    \resizebox{0.7\columnwidth}{!}{
    \begin{tabular}{c|cc}
        \Xhline{2.0\arrayrulewidth}
        Mask ratio & mIoU & mAcc \\
        \Xhline{2.0\arrayrulewidth}
        MinkUNet & 68.5 & 75.2\\
        \Xhline{2.0\arrayrulewidth}
        0\% & 69.3 (\textcolor{red}{+0.8}) & 75.4 (\textcolor{red}{+0.2})\\
        25\% & 69.7 (\textcolor{red}{+1.2}) & 76.0 (\textcolor{red}{+0.8}) \\
        \textbf{50\%} & \textbf{70.1 (\textcolor{red}{+1.6})} & \textbf{76.3 (\textcolor{red}{+1.1})} \\
        75\% & 69.6 (\textcolor{red}{+1.1}) & 75.6 (\textcolor{red}{+0.4}) \\
        90\% & 68.9 (\textcolor{red}{+0.4}) & 75.0 (\textcolor{blue}{-0.2}) \\
        \Xhline{2.0\arrayrulewidth}
    \end{tabular}}
    \caption{Ablation study on mask ratio. 3D semantic segmentation mIoU and mAcc on S3DIS \textit{Area-5}.}
    \label{table7}
\end{table}

\textbf{Rending targets.} Common neural rendering targets include RGB color images and depth images. In this work, we only use RGB color images as pre-training supervision. We conduct an ablation experiment to study the influence of different rendering targets with the transferring task of 3D semantic segmentation. As shown in Table \ref{table8}, using both RGB color and depth images as pre-training supervision does not obtain remarkable performance gains compared to using only RGB color images. In addition, adding additional depth images as supervision also increases the pre-training time and memory consumption.

\begin{table}[h]
    \centering
    \resizebox{0.75\columnwidth}{!}{
    \begin{tabular}{l|cc}
        \Xhline{2.0\arrayrulewidth}
        Supervision & mIoU & mAcc \\
        \Xhline{2.0\arrayrulewidth}
        MinkUNet & 68.5 & 75.2\\
        \Xhline{2.0\arrayrulewidth}
        \textbf{+ Color} & \textbf{70.1 (\textcolor{red}{+1.6})} & \textbf{76.3 (\textcolor{red}{+1.1})} \\
        + Color + Depth & 70.3 (\textcolor{red}{+1.8}) & 76.0 (\textcolor{red}{+0.8}) \\
        \Xhline{2.0\arrayrulewidth}
    \end{tabular}}
    \caption{Ablation study on supervision type. 3D semantic segmentation mIoU and mAcc on S3DIS \textit{Area-5}.}
    \label{table8}
\end{table}

\textbf{Number of input views.} During pre-training, our method use sparse-view RGB images with overlapping regions to produce scene Gaussians for image rendering. In this ablation experiment, we explore the influence of the number of input views on the downstream segmentation task. Table \ref{table9} lists the segmentation results of our fine-tuned model with different number of input views. We observe that the best results are achieved when the number of input views is 3. This is mainly because that, more input views can help our GS$^3$ achieve better rendering quality and thus obtain more accurate supervision from 2D images. However, this significantly increase pre-training time and memory consumption of our GS$^3$ framework. Consequently, we set the number of input views to 2 to balance pre-training overhead and performance.

\begin{table}[t]
    \centering
    \resizebox{0.7\columnwidth}{!}{
    \begin{tabular}{c|cc}
        \Xhline{2.0\arrayrulewidth}
         \#View & mIoU & mAcc \\
        \Xhline{2.0\arrayrulewidth}
        MinkUNet & 68.5 & 75.2\\
        \Xhline{2.0\arrayrulewidth}
         \textbf{2} & \textbf{70.1 (\textcolor{red}{+1.6})} & \textbf{76.3 (\textcolor{red}{+1.1})} \\
         3 & 70.5 (\textcolor{red}{+2.0}) & 76.8 (\textcolor{red}{+1.6}) \\
        \Xhline{2.0\arrayrulewidth}
    \end{tabular}}
    \caption{Ablation study on the number of input views. 3D semantic segmentation mIoU and mAcc on S3DIS.}
    \label{table9}
\end{table}

\textbf{Input image resolution.} In our method, GS$^3$ back-projects the input images into 3D space for self-supervised learning of the point cloud encoder. Higher image resolution can not only provide more detailed 2D image supervision, but also obtain point cloud data with more geometric information. In this ablation experiment, we investigate the effect of input image resolution on our fine-tuned model. As shown in Table \ref{table10}, higher image resolution indeed leads to greater performance improvements. However, this also inevitably increases the overhead of the pre-training process. Therefore, we choose the input image resolution as 320 $\times$ 240 to achieve the best trade-off between pre-training overhead and performance.

\begin{table}[h]
    \centering
    \resizebox{0.7\columnwidth}{!}{
    \begin{tabular}{c|cc}
        \Xhline{2.0\arrayrulewidth}
        Resolution & mIoU & mAcc \\
        \Xhline{2.0\arrayrulewidth}
        MinkUNet & 68.5 & 75.2 \\
        \Xhline{2.0\arrayrulewidth} 
         256 $\times$ 192 & 69.7 (\textcolor{red}{+1.2}) & 76.0 (\textcolor{red}{+0.8}) \\
         \textbf{320 $\times$ 240} & \textbf{70.1 (\textcolor{red}{+1.6})} & \textbf{76.3 (\textcolor{red}{+1.1})} \\
         512 $\times$ 384 & 70.5 (\textcolor{red}{+2.0}) & 76.4 (\textcolor{red}{+1.2}) \\
        \Xhline{2.0\arrayrulewidth}
    \end{tabular}}
    \caption{Ablation study on input image resolution. 3D semantic segmentation mIoU and mAcc on S3DIS \textit{Area-5}. Input image resolution is in the form of width $\times$ height.}
    \label{table10}
\end{table}
\section{Conclusion}
\label{sec:conclusion}

In this paper, we propose a 3D Gaussian Splatting-based Self-Supervised (GS$^3$) framework for point cloud representation learning. We utilizes 3D Gaussian Splatting-based neural rendering as the pretext task, which predicts scene 3D Gaussians from learned point cloud features and then uses a tile-based rasterizer for image rendering. Compared to existing rendering-based frameworks, our method achieves significant pre-training speedup and requires considerably less memory. The point cloud encoder pre-trained by our framework can be well transferred to various downstream tasks. Consistent improvements on four down-stream tasks demonstrate the strong transferability of the point cloud encoder.

In the future, several directions can be explored. First, the recent advances in 3D Gaussian Splatting help our GS$^3$ obtain high-quality rendered images, thereby enhancing the transferability of the point cloud encoder. Second, our GS$^3$ framework can be extended to the 2D image domain.

%and transferred to other downstream tasks.

{
    \small
    \bibliographystyle{ieeenat_fullname}

}

% WARNING: do not forget to delete the supplementary pages from your submission 
\appendix
\renewcommand{\thesection}{A.\arabic{section}}
\renewcommand{\thesubsection}{A.\arabic{section}.\arabic{subsection}}

\section{Visualization of the SR-UNet and PointNet++ Encoder}

The proposed GS$^3$ framework is capable of accommodating various point cloud encoders, including point-based and discretization-based. In this paper, we use PoinbtNet++ (point-based) and SR-UNet (discretization-based) as our encoders for both pre-training and fine-tuning. 

\begin{figure*}
    \centering
    \includegraphics[width=0.8\linewidth]{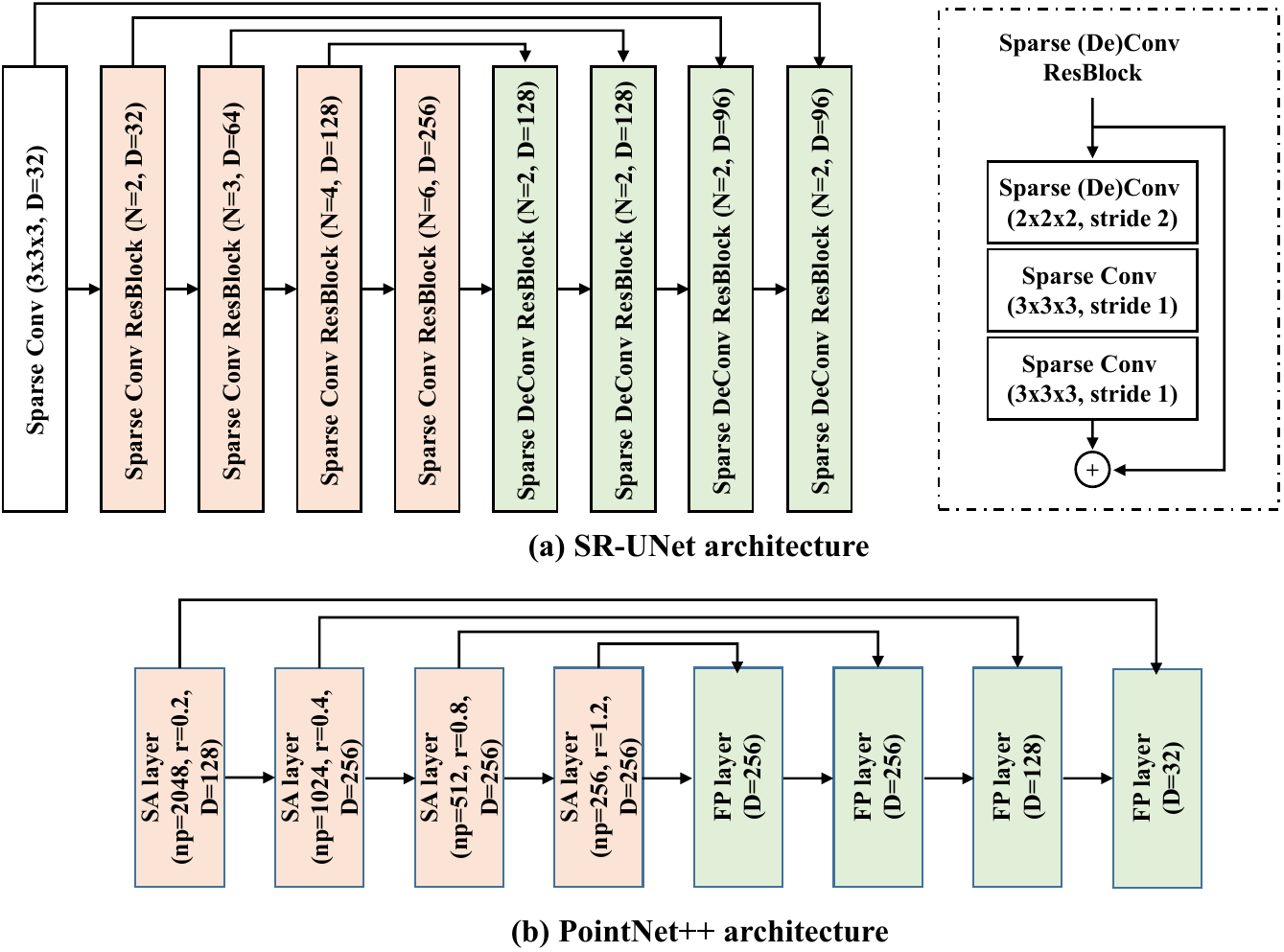}
    \caption{The network architecture of our feature encoder. (a) SR-UNet and (b) PointNet++. For SR-UNet, each sparse (de)convolution layer is followed by a batch norm (BN) layer and a ReLU activation layer. D is the output dimension and N is the number of repeated layers. For PointNet++, SA represents the set abstraction layer, while FP denotes the feature propogation layer. np and r represent the number of down-sampling points and radiu for each SA layer. }
    \label{supp_fig_1}
\end{figure*}

We first introduce the SR-UNet architecture, as shown in Fig. \ref{supp_fig_1}(a). SR-UNet follows the classic UNet encoder-decoder segmentation framework and is mainly implemented by Sparse Convolution (SpConv) and Sparse Deconvolution (SpDeconv). The encoder network consists of five SpConv blocks, and the decoder network has four SpDeconv blocks. Each Spconv / SpDeconv block follows the 2D ResNet basic block design, i.e., each convolution / deconvolution layer is followed by a batch normalization (BN) layer and a ReLU activation layer.

The visualization of the PointNet++ architecture is shown in Fig. \ref{supp_fig_1}(b). PointNet++ consists of four set abstraction (SA) layers and four feature propagation (FP) layers. The number of down-sampling points and radii of these four SA layers are $[2048, 1024, 512, 256]$ and $[0.2, 0.4, 0.8, 1.2]$, respectively.

\section{More Experimental Results}
In this section, we provide detailed experimental results for downstream tasks.

\textbf{More 3D object detection results.} Table \ref{supp_table_1} presents the average precision (AP) values of each category on the SUN RGB-D. We find that our GS$^3$ framework can significantly improve the overall detection performance of the baseline VoteNet, increasing mAP@0.5 by 3.0\% for SUN-RGBD. We also observe that, our GS$^3$ improve the baseline VoteNet in 8 out of 10 categories on the SUN RGB-D dataset.

\textbf{More 3D semantic segmentation results.} Table \ref{supp_table_2} and Table \ref{supp_table_3} list the mean IoU (mIoU) values of each category on the S3DIS and ScanNet v2 datasets, respectively. We note that, with a voxel size of 2cm, MinkUNet pre-trained with our GS$^3$ obtains significant gains in 9 out of 13 categories on the S3DIS dataset (Area5-test), and in 18 out of 20 categories on the ScanNet v2 dataset. Similar improvements are also observed with a voxel size of 5cm. 

\textbf{More 3D instance segmentation results.} Table \ref{supp_table_4} and Table \ref{supp_table_5} report the AP@0.5 values of each category on the S3DIS and ScanNet v2 datasets, respectively. Remarkable improvements for most of semantic categories are observed on both S3DIS and ScanNet v2 dataset.

\begin{figure*}[t]
    \centering
    \includegraphics[width=\linewidth]{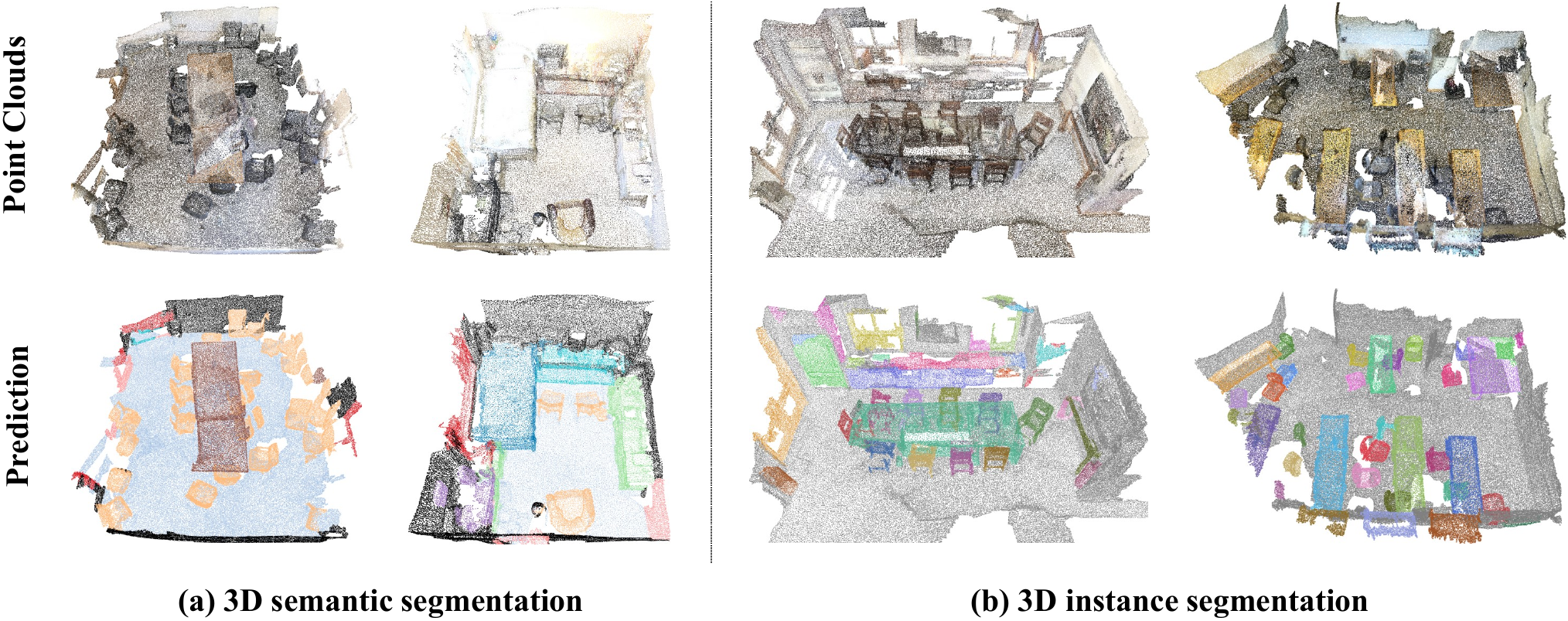}
    \caption{Qualitative results of our fine-tuned model on downstream (a) 3D semantic segmentation and (b) 3D instance segmentation tasks.}
    \label{supp_fig_2}
\end{figure*}

\textbf{Qualitative results.} Figure \ref{supp_fig_2} shows the visualization results of our fine-tuned model on downstream 3D semantic segmentation and 3D instance segmentation tasks.

\begin{table*}[h]
    \centering
    \resizebox{\textwidth}{!}{
    \begin{tabular}{c|c|cccccccccc}
        \Xhline{2.0\arrayrulewidth}
        Method & mAP@0.5 & bathtub & bed & bookshelf & chair & desk & dresser & nightstand & sofa & table & toilet \\
        \Xhline{2.0\arrayrulewidth}
        VoteNet \cite{VoteNet} & 33.7 & 47.0 & 50.1 & 7.2 & 53.9 & 5.3 & 11.5 & 40.7 & 42.4 & 19.5 & 59.8 \\
        \multirow{2}{*}{\textbf{GS$^3$} + VoteNet} & 36.7 & 54.7 & 53.0 & 10.0 & 53.9 & 7.5 & 17.8 & 40.3 & 51.1 & 17.6 & 61.1 \\
        & \textcolor{red}{(+3.0)} & \textcolor{red}{(+7.7)} & \textcolor{red}{(+2.9)} & \textcolor{red}{(+2.8)} & \textcolor{red}{(+0.0)} & \textcolor{red}{(+2.2)} & \textcolor{red}{(+6.3)} & \textcolor{blue}{(-0.4)} & \textcolor{red}{(+8.7)} & \textcolor{blue}{(-1.9)} & \textcolor{red}{(+1.3)} \\
        \Xhline{2.0\arrayrulewidth}
    \end{tabular}}
    \caption{Comparative 3D object detection results for each category on the SUN-RGBD dataset, evaluated with mAP@0.5. The number in each bracket denotes the performance improvement (shown in red) or degradation (shown in blue) compared to the corresponding baseline.}
    \label{supp_table_1}
\end{table*}

\begin{table*}[t]
    \centering
    \resizebox{\textwidth}{!}{
    \begin{tabular}{r|cc|ccccccccccccc}
        \Xhline{2.0\arrayrulewidth}
        Method & mIoU & mAcc & ceil. & floor & wall & beam & col. & wind. & door & chair & table & book. & sofa & board & clut. \\
        \Xhline{2.0\arrayrulewidth}
        PointNet \cite{PointNet} & 41.1 & 49.0 & 88.8 & 97.3 & 69.8 & 0.1 & 3.9 & 46.3 & 10.8 & 52.6 & 58.9 & 40.3 & 5.9 & 26.4 & 33.2 \\
        %PointWeb \cite{PointWeb} & 60.3 & 66.6 & 92.0 & 98.5 & 79.4 & 0.0 & 21.1 & 59.7 & 34.8 & 88.3 & 76.3 & 69.3 & 46.9 & 64.9 & 52.5 \\
        KPConv \cite{KPConv} & 67.1 & 72.8 & 92.8 & 97.3 & 82.4 & 0.0 & 23.9 & 58.0 & 69.0 & 91.0 & 81.5 & 75.3 & 75.4 & 66.7 & 58.9 \\
        MinkUNet (5cm) \cite{MinkUNet} & 65.4 & 71.7 & 91.8 & 98.7 & 86.2 & 0.0 & 34.1 & 48.9 & 62.4 & 89.8 & 81.6 & 74.9 & 47.2 & 74.4 & 58.6 \\
        Point Transformer \cite{PT} & 70.4 & 76.5 & 94.0 & 98.5 & 86.3 & 0.0 & 38.0 & 63.4 & 74.3 & 82.4 & 89.1 & 80.2 & 74.3 & 76.0 & 59.3 \\
        \Xhline{2.0\arrayrulewidth}    
        MinkUNet$^\dagger$ (5cm) \cite{MinkUNet} & 62.8 & 70.6 & 90.8 & 96.1 & 81.4 & 0.1 & 18.8 & 53.3 & 60.7 & 84.9 & 75.8 & 69.1 & 61.8 & 68.5 & 55.0 \\
        \multirow{2}{*}{\textbf{GS$^3$} + MinkUNet (5cm)} & 63.8 & 71.3 & 91.4 & 96.7 & 81.6 & 0.1 & 25.5 & 53.5 & 55.5 & 86.3 & 75.2 & 69.3 & 66.6 & 73.3 & 54.4 \\
        & \textcolor{red}{(+1.0)} & \textcolor{red}{(+0.7)} & \textcolor{red}{(+0.6)} & \textcolor{red}{(+0.6)} & \textcolor{red}{(+0.2)} & \textcolor{red}{(+0.0)} & \textcolor{red}{(+6.7)} & \textcolor{red}{(+0.2)} & \textcolor{blue}{(-5.2)} & \textcolor{red}{(+1.4)} & \textcolor{blue}{(-0.6)} & \textcolor{red}{(+0.2)} & \textcolor{red}{(+4.8)} & \textcolor{red}{(+4.8)} & \textcolor{blue}{(-0.6)} \\
        MinkUNet$^\dagger$ (2cm) \cite{MinkUNet} & 68.5 & 75.2 & 91.6 & 97.6 & 84.1 & 0.0 & 24.5 & 60.3 & 77.5 & 87.8 & 81.6 & 72.6 & 73.8 & 80.3 & 59.0 \\
        \multirow{2}{*}{\textbf{GS$^3$} + MinkUNet (2cm)} & 70.1 & 76.3 & 92.7 & 97.9 & 84.5 & 0.1 & 34.7 & 63.2 & 78.5 & 89.8 & 81.8 & 72.2 & 76.1 & 80.0 & 59.3 \\
        & \textcolor{red}{(+1.6)} & \textcolor{red}{(+1.1)} & \textcolor{red}{(+1.1)} & \textcolor{red}{(+0.3)} & \textcolor{red}{(+0.4)} & \textcolor{red}{(+0.0)} & \textcolor{red}{(+10.2)} & \textcolor{red}{(+2.9)} & \textcolor{red}{(+1.0)} & \textcolor{red}{(+2.0)} & \textcolor{red}{(+0.2)} & \textcolor{blue}{(-0.4)} & \textcolor{red}{(+2.3)} & \textcolor{blue}{(-0.3)} & \textcolor{red}{(+0.3)} \\
        \Xhline{2.0\arrayrulewidth}
    \end{tabular}}
    \caption{Comparative 3D semantic segmentation results for each category on the S3DIS (Area-5) dataset. $\dagger$ denotes the reproduced results. The number in each bracket denotes the performance improvement (shown in red) or degradation (shown in blue) compared to the corresponding baseline.}
    \label{supp_table_2}
\end{table*}

\begin{table*}[t]
    \centering
    \resizebox{\textwidth}{!}{
    \begin{tabular}{r|cc|cccccccccccccccccccc}
        \Xhline{2.0\arrayrulewidth}
        {Method} & \rotatebox{90}{mIoU} & \rotatebox{90}{mAcc} & \rotatebox{90}{wall} & \rotatebox{90}{floor} & \rotatebox{90}{cabinet} & \rotatebox{90}{bed} & \rotatebox{90}{chair} & \rotatebox{90}{sofa} & \rotatebox{90}{table} & \rotatebox{90}{door} & \rotatebox{90}{window} & \rotatebox{90}{bookshelf} & \rotatebox{90}{picture} & \rotatebox{90}{counter} & \rotatebox{90}{desk} & \rotatebox{90}{curtain} & \rotatebox{90}{refrigerator} & \rotatebox{90}{shower curtain} & \rotatebox{90}{toilet} & \rotatebox{90}{sink} & \rotatebox{90}{bathtub} & \rotatebox{90}{other furniture} \\
        \Xhline{2.0\arrayrulewidth}    
        MinkUNet$^\dagger$ (5cm) \cite{MinkUNet} & 66.6 & 75.0 & 81.1 & 95.3 & 63.6 & 81.3 & 88.3 & 83.5 & 74.0 & 53.0 & 56.1 & 71.7 & 21.0 & 59.4 & 63.3 & 50.5 & 43.0 & 58.3 & 89.6 & 61.6 & 85.2 & 51.5 \\
        \multirow{2}{*}{\textbf{GS$^3$} + MinkUNet (5cm)} & 68.2 & 76.4 & 82.3 & 95.8 & 64.5 & 79.5 & 89.1 & 86.0 & 74.8 & 56.4 & 56.0 & 75.2 & 23.9 & 59.5 & 62.4 & 56.4 & 45.8 & 61.9 & 92.5 & 61.7 & 86.5 & 53.5 \\
        & \textcolor{red}{(+1.6)} & \textcolor{red}{(+1.4)} & \textcolor{red}{(+1.2)} & \textcolor{red}{(+0.5)} & \textcolor{red}{(+0.9)} & \textcolor{blue}{(-1.8)} & \textcolor{red}{(+0.8)} & \textcolor{red}{(+2.5)} & \textcolor{red}{(+0.8)} & \textcolor{red}{(+3.4)} & \textcolor{blue}{(-0.1)} & \textcolor{red}{(+3.5)} & \textcolor{red}{(+2.9)} & \textcolor{red}{(+0.1)} & \textcolor{blue}{(-0.9)} & \textcolor{red}{(+5.9)} & \textcolor{red}{(+2.8)} & \textcolor{red}{(+3.6)} & \textcolor{red}{(+2.9)} & \textcolor{red}{(+0.1)} & \textcolor{red}{(+1.3)} & \textcolor{red}{(+2.0)} \\
        MinkUNet$^\dagger$ (2cm) \cite{MinkUNet} & 71.9 & 80.6 & 85.8 & 96.3 & 65.7 & 79.5 & 89.9 & 84.5 & 71.3 & 65.4 & 60.3 & 79.4 & 35.3 & 64.9 & 63.0 & 73.0 & 54.5 & 68.0 & 93.1 & 66.3 & 85.2 & 57.0\\
        \multirow{2}{*}{\textbf{GS$^3$} + MinkUNet (2cm)} & 73.4 & 81.0 & 85.9 & 96.5 & 66.9 & 81.6 & 91.6 & 86.7 & 75.6 & 66.4 & 61.2 & 82.5 & 30.5 & 63.7 & 67.5 & 76.3 & 57.7 & 69.3 & 93.2 & 66.7 & 87.4 & 60.2 \\
        & \textcolor{red}{(+1.5)} & \textcolor{red}{(+0.4)} & \textcolor{red}{(+0.1)} & \textcolor{red}{(+0.2)} & \textcolor{red}{(+1.2)} & \textcolor{red}{(+2.1)} & \textcolor{red}{(+1.7)} & \textcolor{red}{(+2.2)} & \textcolor{red}{(+4.3)} & \textcolor{red}{(+1.0)} & \textcolor{red}{(+0.9)} & \textcolor{red}{(+3.1)} & \textcolor{blue}{(-4.8)} & \textcolor{blue}{(-1.2)} & \textcolor{red}{(+4.5)} & \textcolor{red}{(+3.3)} & \textcolor{red}{(+3.2)} & \textcolor{red}{(+1.3)} & \textcolor{red}{(+0.1)} & \textcolor{red}{(+0.4)} & \textcolor{red}{(+2.2)} & \textcolor{red}{(+3.2)} \\
        \Xhline{2.0\arrayrulewidth}   
    \end{tabular}}
    \caption{Comparative 3D semantic segmentation results for each category on the ScanNet v2 val set. $\dagger$ denotes the reproduced results. The number in each bracket denotes the performance improvement (shown in red) or degradation (shown in blue) compared to the corresponding baseline.}
    \label{supp_table_3}
\end{table*}

\begin{table*}[h]
    \centering
    \resizebox{\textwidth}{!}{
    \begin{tabular}{r|c|cccccccccccc}
        \Xhline{2.0\arrayrulewidth}
        Method & AP@50 & ceil. & floor & wall & beam & col. & wind. & door & chair & table & book. & sofa & board \\
        \Xhline{2.0\arrayrulewidth} 
        PointGroup$^\dagger$ (5cm) \cite{PointGroup} & 55.7 & 46.2 & 95.5 & 64.0 & 0.0 & 37.1 & 72.1 & 55.0 & 64.3 & 29.6 & 35.4 & 88.4 & 80.6 \\
        \multirow{2}{*}{\textbf{GS$^3$} + PointGroup (5cm)} & 57.7 & 45.7 & 96.9 & 64.5 & 0.0 & 39.0 & 61.8 & 72.7 & 63.5 & 42.9 & 31.2 & 90.0 & 84.7\\
        & \textcolor{red}{(+2.0)} & \textcolor{blue}{(-0.5)} & \textcolor{red}{(+1.4)} & \textcolor{red}{(+0.5)} & \textcolor{red}{(+0.0)} & \textcolor{red}{(+1.9)} & \textcolor{blue}{(-10.3)} & \textcolor{red}{(+17.7)} & \textcolor{blue}{(-0.8)} & \textcolor{red}{(+13.3)} & \textcolor{blue}{(-4.2)} & \textcolor{red}{(+1.6)} & \textcolor{red}{(+4.1)} \\
        PointGroup$^\dagger$ (2cm) \cite{PointGroup} & 59.4 & 67.9 & 99.9 & 67.5 & 0.0 & 38.0 & 68.5 & 85.2 & 93.9 & 31.0 & 25.1 & 53.1 & 82.3\\
        \multirow{2}{*}{\textbf{GS$^3$} + PointGroup (2cm)} & 61.1 & 55.8 & 97.5 & 60.0 & 0.0 & 47.9 & 76.9 & 70.3 & 91.0 & 33.2 & 33.4 & 81.8 & 85.7 \\
        & \textcolor{red}{(+1.7)} & \textcolor{blue}{(-12.1)} & \textcolor{blue}{(-2.4)} & \textcolor{blue}{(-7.5)} & \textcolor{red}{(+0.0)} & \textcolor{red}{(+9.9)} & \textcolor{red}{(+8.4)} & \textcolor{blue}{(-14.9)} & \textcolor{blue}{(-2.9)} & \textcolor{red}{(+2.2)} & \textcolor{red}{(+8.3)} & \textcolor{red}{(+28.7)} & \textcolor{red}{(+3.4)} \\
        \Xhline{2.0\arrayrulewidth} 
    \end{tabular}}
    \caption{Comparative 3D instance segmentation results for each category on the S3DIS Area-5 set. $\dagger$ denotes the reproduced results. The number in each bracket denotes the performance improvement (shown in red) or degradation (shown in blue) compared to the corresponding baseline.}
    \label{supp_table_4}
\end{table*}

\begin{table*}[h]
    \centering
    \resizebox{\textwidth}{!}{
    \begin{tabular}{r|c|cccccccccccccccccc}
        \Xhline{2.0\arrayrulewidth}
        {Method} & \rotatebox{90}{AP@50} & \rotatebox{90}{cabinet} & \rotatebox{90}{bed} & \rotatebox{90}{chair} & \rotatebox{90}{sofa} & \rotatebox{90}{table} & \rotatebox{90}{door} & \rotatebox{90}{window} & \rotatebox{90}{bookshelf} & \rotatebox{90}{picture} & \rotatebox{90}{counter} & \rotatebox{90}{desk} & \rotatebox{90}{curtain} & \rotatebox{90}{refrigerator} & \rotatebox{90}{shower curtain} & \rotatebox{90}{toilet} & \rotatebox{90}{sink} & \rotatebox{90}{bathtub} & \rotatebox{90}{other furniture} \\
        \Xhline{2.0\arrayrulewidth} 
        PointGroup$^\dagger$ (5cm) \cite{PointGroup} & 49.1 & 48.5 & 70.3 & 77.0 & 64.9 & 66.2 & 38.3 & 24.8 & 45.3 & 15.1 & 25.5 & 30.2 & 27.5 & 54.9 & 54.2 & 94.8 & 39.8 & 76.9 & 29.7\\
        \multirow{2}{*}{\textbf{GS$^3$} + PointGroup (5cm)} & 50.6 & 47.7 & 73.0 & 78.6 & 67.6 & 68.8 & 39.3 & 30.2 & 48.3 & 17.2 & 21.8 & 29.9 & 19.3 & 56.9 & 55.8 & 95.1 & 48.2 & 77.7 & 35.1 \\
        & \textcolor{red}{(+1.5)} & \textcolor{blue}{(-0.8)} & \textcolor{red}{(+2.7)} & \textcolor{red}{(+1.6)} & \textcolor{red}{(+2.7)} & \textcolor{red}{(+2.6)} & \textcolor{red}{(+1.0)} & \textcolor{red}{(+5.4)} & \textcolor{red}{(+3.0)} & \textcolor{red}{(+2.1)} & \textcolor{blue}{(-3.7)} & \textcolor{blue}{(-0.3)} & \textcolor{blue}{(-8.2)} & \textcolor{red}{(+2.0)} & \textcolor{red}{(+1.6)} & \textcolor{red}{(+0.3)} & \textcolor{red}{(+8.4)} & \textcolor{red}{(+0.8)} & \textcolor{red}{(+5.4)}\\
        PointGroup$^\dagger$ (2cm) \cite{PointGroup} & 57.6 & 49.9 & 72.5 & 87.1 & 59.6 & 67.2 & 48.5 & 38.7 & 61.2 & 32.0 & 21.8 & 28.5 & 43.6 & 54.4 & 70.0 & 98.3 & 69.4 & 79.4 & 54.7\\
        \multirow{2}{*}{\textbf{GS$^3$} + PointGroup (2cm)} & 59.2 & 53.5 & 74.1 & 88.9 & 72.0 & 69.0 & 47.2 & 36.3 & 54.7 & 34.5 & 27.2 & 29.5 & 46.6 & 64.7 & 66.8 & 99.9 & 67.6 & 77.4 & 56.2 \\
        & \textcolor{red}{(+1.6)} & \textcolor{red}{(+3.6)} & \textcolor{red}{(+1.6)} & \textcolor{red}{(+1.8)} & \textcolor{red}{(+12.4)} & \textcolor{red}{(+1.8)} & \textcolor{blue}{(-1.3)} & \textcolor{blue}{(-2.4)} & \textcolor{blue}{(-6.5)} & \textcolor{red}{(+2.5)} & \textcolor{red}{(+5.4)} & \textcolor{red}{(+1.0)} & \textcolor{red}{(+3.0)} & \textcolor{red}{(+10.3)} & \textcolor{blue}{(-3.2)} & \textcolor{red}{(+1.6)} & \textcolor{blue}{(-1.8)} & \textcolor{blue}{(-2.0)} & \textcolor{red}{(+1.5)} \\
        \Xhline{2.0\arrayrulewidth} 
    \end{tabular}}
    \caption{Comparative 3D instance segmentation results for each category on the ScanNet v2 val set. $\dagger$ denotes the reproduced results. The number in each bracket denotes the performance improvement (shown in red) or degradation (shown in blue) compared to the corresponding baseline.}
    \label{supp_table_5}
\end{table*}

\end{document}